\documentclass[letterpaper, 10 pt, conference]{ieeeconf}  

\IEEEoverridecommandlockouts                              
\usepackage{amsmath, amssymb, amsfonts}
\usepackage{algorithm, algorithmic}
\usepackage{graphicx}
    \graphicspath{{figures/}}
\usepackage{mathtools}
\usepackage{textcomp}
\usepackage{xcolor}
\def\BibTeX{{\rm B\kern-.05em{\sc i\kern-.025em b}\kern-.08em
    T\kern-.1667em\lower.7ex\hbox{E}\kern-.125emX}}
\usepackage{caption}
\usepackage{subcaption}

\newcommand{\probP}{\text{I\kern-0.15em P}}
\newcommand{\probE}{\text{I\kern-0.15em E}}

\title{\LARGE \bf
Developing Driving Strategies Efficiently: A Skill-Based Hierarchical Reinforcement Learning Approach
\\
}

\author{Yigit Gurses, Kaan Buyukdemirci, and Yildiray Yildiz
\thanks{Y. Gurses is with Department of Computer Engineering,
    Email:{ \tt\small yigit.gurses@ug.bilkent.edu.tr},
K. Buyukdemirci is with the Department of Electrical and Electronical Engineering,
    Email:{ \tt\small kaan.buyukdemirci@ug.bilkent.edu.tr},
 Y. Yildiz is with the Department of Mechanical Engineering,
    Email:{ \tt\small yyildiz@bilkent.edu.tr},
    Bilkent University 06800 Bilkent, Ankara, Turkey.}
}

\begin{document}

\maketitle
\thispagestyle{empty}
\pagestyle{empty}

\begin{abstract}

Driving in dense traffic with human and autonomous drivers is a challenging task that requires high-level planning and reasoning. Human drivers can achieve this task comfortably, and there has been many efforts to model human driver strategies. These strategies can be used as inspirations for developing autonomous driving algorithms or to create high-fidelity simulators. Reinforcement learning is a common tool to model driver policies, but conventional training of these models can be computationally expensive and time-consuming. To address this issue, in this paper, we propose ``skill-based" hierarchical driving strategies, where motion primitives, i.e. skills, are designed and used as high-level actions. This reduces the training time for applications that require multiple models with varying behavior. Simulation results in a merging scenario demonstrate that the proposed approach yields driver models that achieve higher performance with less training compared to baseline reinforcement learning methods.

\end{abstract}

\section{INTRODUCTION}

Deep reinforcement learning (RL) has solved many challenging problems \cite{Dota, Atari} and achieved super-human performance in complex tasks that classical algorithms struggle with \cite{Silver Go, Silver All}. Autonomous driving could be considered such a problem, and to that extent, there is growing literature on utilising deep RL in this field \cite{RL AV 1, RL AV 2, RL AV 3, RL AV 4, RL AV 5}. A comprehensive review of recent developments on this subject can be found in \cite{RL AV Survey}.

In a traffic environment, there are critical events such as collisions that occur rarely. Although rare, they are valuable for the learning process, but they lead to environments with sparse rewards. RL methods are known to suffer from extensive training times and sub-optimal long-term planning when dealing with sparse rewards. One way to solve this problem is incorporating human experts' domain knowledge through reward shaping \cite{Reward Shaping First, Reward Shaping Invariance}. However, this introduces human bias to the process and can lead to sub-optimal performance and unwanted behavior. 

Hierarchical RL (HRL) is a branch of RL that can be helpful in solving the sparse rewards problem, where a set of policies are employed as actions of a high-level agent \cite{Macro HRL, Feudal HRL, Option-Critic, Act to Reason, Kickstarting, Hierarchical Kickstarting, HRL Book, HRL Survey}. This can be useful in overcoming the difficulties posed by sparse rewards in traffic, since low-level policies can be trained using intrinsic dense rewards that are not dependent on the high-level goals. Another point to consider is how the structure of HRL is more similar to human decision making compared to baseline RL. When driving, humans consciously plan their trajectories (similar to high-level agent) and the execution is done by subconscious motor skills (similar to low-level policies) like turning the wheel or hitting the brakes. For this reason, HRL could have the potential to model or interact with humans more successfully than its counterparts.

Low-level policies used in HRL can be obtained using several methods including unsupervised exploration \cite{Active Pre-Training, Marginal Matching, Curiosity, Proto RL}. In this paper, we propose designing motion primitives as ``driving skills" and using them as low-level policies In HRL, in a data-based highway merging scenario. These primitives can be defined as action sequences that form meaningful behaviors, such as ``merging and then accelerating", or ``overtaking on the left". We design these skills using unsupervised skill discovery (USD) \cite{DADS, DIAYN}, which makes them reusable in RL tasks with different reward functions. This provides a scalable approach for obtaining multiple models. We also propose a controlled randomization-based method together with observation binning to obtain diverse and high quality driving skills. 

To summarize, the main contributions of this study are the following.
\begin{enumerate}

\item Creating \textit{driving skills} in a highway merging environment, where the environment is constructed using real traffic data. 
\item Introducing a novel method to improve skill diversity and quality. 
\item Developing driving policies using HRL and driving skills, resulting in significantly enhanced convergence rates and superior performance compared to driving policies obtained using traditional RL methods. 

\end{enumerate}

\section{Method}
In this section, skill-based hierarchical reinforcement learning is explained, starting from the basic building blocks.

\subsection{Markov Decision Process}
A Markov Decision Process (MDP) is a tuple $(S, A, P, r)$, where, $S$ and $A$ are the sets of states and actions, respectively. $P:S \times A \times S \rightarrow [0,1]$ is the function, where $P(s,a,s')$ represents the transition probability from state $s$ to $s'$ if action $a$ is taken. $r:S \times A \rightarrow \mathbb{R}$ is the function where $r(s, a)$ is the reward for taking action $a$ in state $s$.

\subsection{Reinforcement Learning}
In RL, an agent's behavior is represented in a policy function $\pi: S \times A \rightarrow [0,1]$ where $\pi(s,a)$ represents the probability of the agent taking action $a$ in state $s$. In an MDP, a tuple $(s_t,a_t,r_t,s_{t+1})$ is defined as an experience sample and a sequence of experience samples is defined as a trajectory. Given a state-action pair $(s_t,a_t)$, the expected cumulative reward over all possible trajectories following $(s_t,a_t)$ is calculated as
\begin{equation}
    Q^{\pi}(s_t,a_t)=r(s_t,a_t) + \mathbb{E}_{a \sim \pi(s)}\left[ \sum_{i=1}^{\infty}\gamma^{i}r(s_{t+i},a_{t+i})|s_t,a_t \right],
\end{equation}
where $\gamma \in [0,1)$ is the discount factor. The goal of reinforcement learning is to find an optimal policy $\pi^*$ such that $\pi^*=\arg \max_{\pi} Q^{\pi}(s,a), \forall s \in S, \forall a \in A$. $\pi^*$ can be estimated using methods like Q-learning and deep Q-networks (DQN) \cite{DQN}.

\subsection{Motion primitives (skills)}
Below, we provide a basic introduction to the skills concept. Further details can be found in \cite{DIAYN}.

\textit{Mutual Information (I)}: measures the average quantity of information gained about one variable by sampling the other variable. Given two random variables X and Y, their marginal distributions p(X) and p(Y), and joint distribution p(X,Y); the mutual information of X and Y is expressed as 
\begin{equation}
    I(X;Y)~=~\sum_{x \in X}\sum_{y \in Y}p(x,y)log \frac{p(x,y)}{p(x)p(y)}.
\end{equation}
\textit{Entropy (H)}: Entropy is Quantifies the randomness of a random variable. Given a random variable X and its probability distribution p(X), entropy of X is defined as 

\begin{equation}
    H(X)=-\sum_{x \in X}p(x)\log p(x).
\end{equation}

The goal of is to obtain task agnostic skill policies, each defining a different skill, that explore different sub-spaces of a state space. The following goals are prioritized.

\begin{itemize}
    \item We want skills to be inferable from given states. Maximizing the mutual information between skills ($Z$) and states ($S$), i.e. maximizing $I(S;Z)$, achieves this goal.
    \item We do not want skills to be inferable from actions since different actions can lead the agent to the same states. This would be counterproductive for the goal of exploring the whole state space. Minimizing mutual information between actions ($A$) and skills given states, i.e. minimizing $I(A;Z|S)$, achieves this goal.
    \item We want each skill to act as randomly as possible and so explore as large of a state space as possible. Maximizing the  entropy ($H$) of actions over states, i.e maximizing $H[A|S]$, achieves this goal. 
\end{itemize}

These goals translate to maximizing the following objective function,
\begin{equation}
\begin{split}
    \mathcal{F}(\theta) & \triangleq I(S;Z) + H[A|S] - I(A;Z|S) \\
                        & = (H[Z] - H[Z;S]) + H[A|S] \\ & \hspace{5mm} - (H[A|S] - H[A|S;Z]) \\
                        & = H[Z] - H[Z;S] + H[A|S;Z],
\end{split}
\end{equation}
 where $\theta$ represents the parameters of the policy $\pi_\theta$. To maximize the first term $H[Z]$, the probability distribution $p(Z)$ is selected as a uniform distribution. $p(z)$ is the probability of the skill $z$ being sampled at the beginning of an episode. To maximize the third term, a soft actor critic (SAC) \cite{SAC} agent that maximizes the entropy of actions $H[A|S;Z]$, while simultaneously maximizing the  expected reward, is trained. Training is done with the reward function
\begin{equation}
    r_z(s,a)= \log q_{\phi}(z|s) - \log p(z),
\end{equation}
where $q_{\phi}(z|s)$ is the output of a discriminator network that is trained concurrently with the SAC agent to predict $p(z|s)$. This reward increases the inferability of skills from states, and therefore minimizes the second term $H[Z;S]$. Once the training is complete, the policy $\pi_{\theta}$, which gives a probability distribution of actions given a state and skill, is obtained. The probability of the agent selecting the action $a$ given a state-skill pair $(s,z)$ is calculated as $\probP(A=a)=\pi_{\theta}(s,a|z)$.
By selecting a constant skill $z$, we can define a policy $\pi_z$ such that $\forall z \in Z; \forall s \in S; \forall a \in A; \pi_z(s,a)=\pi_{\theta}(s,a|z)$. In other words, $\pi_z$ is the policy of skill $z$ induced from $\pi_{\theta}$.

\subsection{Skill-Based Hierarchical Learning} \label{sec:HRL-USD}
In hierarchial reinforcement learning (HRL), the highest level task and a given subset are represented by $\Gamma$ and $\omega$, respectively. $\pi_{\Gamma}$ is defined as the policy that maps the state space to subtasks, and $\pi_{\omega}$ as the policy that maps the state space to low level actions. 

In this study, we define $\Gamma$ as the task of reaching the end of the highway region without any collisions while staying as close as possible to the desired velocity and headway distances. Each skill, represented by z, is considered as a subtask of $\Gamma$ with the corresponding policy $\pi_{z}$, as explained in the previous section.

\section{Traffic Scenario}
We are interested in a merging scenario in a highway environment. To build a high-fidelity simulation environment, we use real traffic-data that is available online. The details are explained in the following subsections.

\subsection{Obtaining Environment Parameters From Real-Life Data} \label{sec:dataparams}

\begin{figure}[t]
  \centering
  \includegraphics[width=0.45\textwidth]{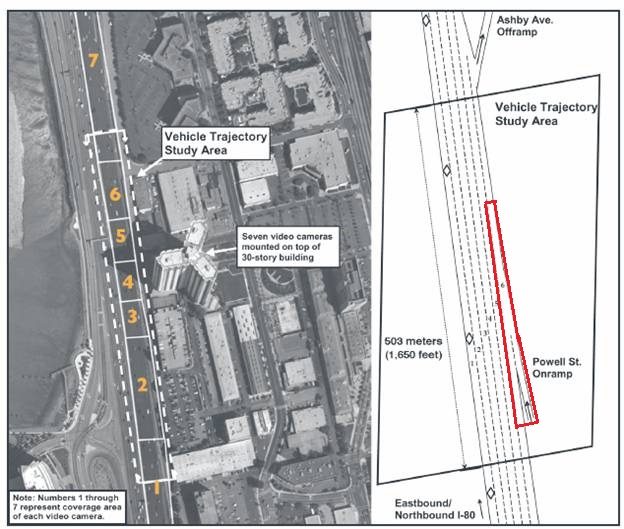}
  \caption{NGSIM I-80 Study Area, Area of Interest Enclosed in Red Lines}
  \label{fig:I80}
\end{figure}
 NGSIM I-80 is a data-set consisting of vehicle trajectories collected on Eastbound I-80 in San Francisco Bay Area at Emeryville, CA, on April 13, 2005 \cite{NGSIM}. The data is collected from a section of the road that is approximately 500 meters, includes six freeway lanes and an on-ramp lane that merges into the freeway (See Fig.~\ref{fig:I80}).

 \begin{figure}[b]
  \centering
  \includegraphics[width=0.5\textwidth]{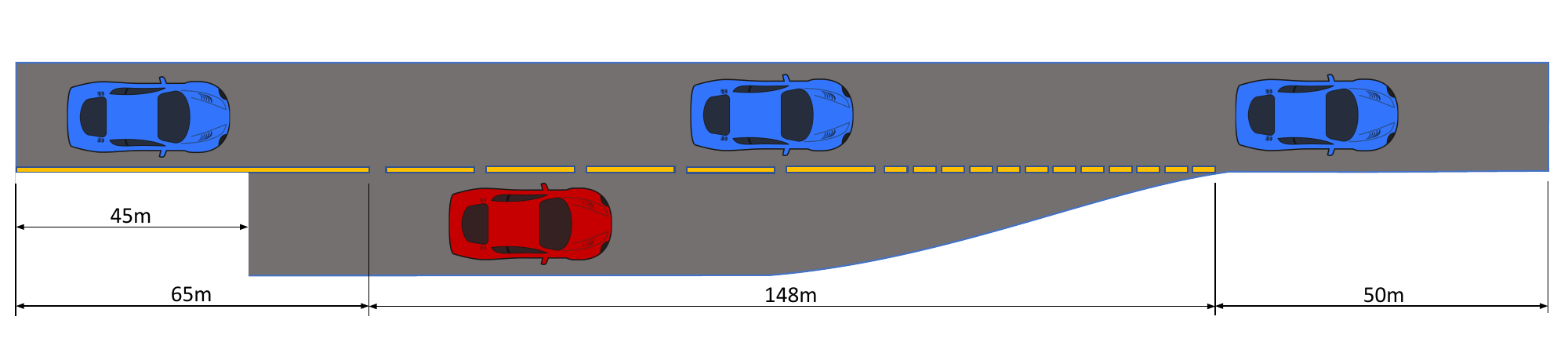}
  \caption{Highway Merging Environment}
  \label{fig:env}
\end{figure}

We use a reconstructed version of NGSIM I-80 \cite{NGSIM Reconstructed} where the lateral vehicle coordinates are represented as discrete lane-numbers, instead of real valued y-coordinates. We process the data collected on the rightmost lane and the ramp to obtain certain statistics to be used in generating the scenario: The mean headway distance and velocity on the road lane is calculated as 23.28m and 9.01m/s, respectively. The dimensions of the road section where the data is collected are used to define the dimensions of the road in the simulated environment.

\subsection{Environment Construction}

The environment consists of a highway lane and an on-ramp lane merging into the highway (see Fig.~\ref{fig:env}). The total length of the road is 263m. The on-ramp ends at 213m, and the legal merging zone starts at 65m, before which the vehicles are not allowed to merge. This results in a 148m-long legal merging zone. All vehicles are assumed to have a length of 5m. At the start of a training episode, 
\begin{itemize}
    \item the ego vehicle starts at the on-ramp at 0 meters with a velocity sampled from a normal distribution with 1 standard deviation and 9.01 mean, and
    \item the highway lane is populated with $n$ cars where each car has a starting velocity sampled from a normal distribution with 1 standard deviation and 9.01 mean. For $i \in \{1, 2, ..., n\}$, $i^{th}$ car starts at the coordinate $x_i=50m(i-1)+x_i^{rand}$ where $x_i^{rand}$ is sampled from a normal distribution with 1 standard deviation and 23.28 mean.
\end{itemize}

\subsection{State Transition Model}

At any time frame $t$, there exists multiple environment vehicles and 1 ego vehicle. Each vehicle has a binary variable $l(t)$ representing their current lanes. $l(t)=0$ implies the vehicle is on the highway lane, and $l(t)=1$ implies the vehicle is on the on-ramp merging lane. Two continuous variables $x(t)$ and $v(t)$ represent x-coordinates and velocities, respectively. At each time frame, vehicles choose an acceleration $a(t)$, and a probability of lane change $l_{p}(t)$. 

The evolution of longitudinal vehicle states is computed as

\begin{equation}
    x(t+\Delta t) = x(t) + v(t) \Delta t + 1/2 a(t) \Delta t ^2,
    \label{eq:xcoord}
\end{equation}
\begin{equation}
    v(t + \Delta t) = v(t) + a(t) \Delta t.
    \label{eq:v}
\end{equation}
For lane change, lane change probability $l_p(t)$ is accepted in the range $[-0.1,1.1]$ and the new lane is calculated as 
\begin{equation}
    l(t + \Delta t) =  \begin{cases} 
      0 & l(t) = 0 \\
      0 & l(t)=1 \text{ \& } l_{p}(t) \geq 1  \\
      0 & l(t)=1 \text{ \& } p < l_{p}(t) \text{ \& } l_{p}(t) > 0 \\
      1 & l(t)=1 \text{ \& } p \geq l_{p}(t) \text{ \& } l_{p}(t) > 0 \\
      1 & l(t)=1 \text{ \& } l_{p}(t) \leq 0 \\
   \end{cases}
\end{equation}
This transition algorithm does not allow vehicles on the highway to change lanes and makes the on-ramp vehicles' lane change probabilistic, instead of deterministic. This allows emerging skills to explore larger state spaces. 

\subsection{Observation Space}
The ego vehicle can observe its own velocity $v_{agent}$, its own lane in one hot representation, a binary variable representing if merging is legal or not, as well as the relative velocities, $v_{rel}$, and distances, $d_{rel}$, of the surrounding vehicles that are in the front, rear, front-left and rear-left. If there is no vehicle to be observed in a possible location, $v_{rel}$ is set to $v_{agent}$, and $d_{rel}$ is set to $d_{max}$.

These observations are normalized in the range $[0,1]$ as

\begin{equation}
    v_{agent}^{norm} = (v_{agent}-v_{min}) / (v_{max}-v_{min})
\end{equation}
\begin{equation}
    v^{norm}_{rel} = (v_{rel}) / (v_{max}-v_{min})
\end{equation}
\begin{equation}
    d^{norm}_{rel} = \begin{cases} 
    d_{rel} / d_{max} & d_{rel} < d_{max} \\
    1 & d_{rel} \geq d_{max},
    \end{cases}
\end{equation}
where $v_{max}$ is the maximum allowed speed, which is set to 29.16m/s, and $d_{max}$ is the maximum observable distance, which is set to 30m. The end of the merging region is treated as a vehicle with zero velocity. 

A real valued state space makes it hard to obtain distinct skills due to the infinitely large space size. To solve this issue and obtain skills as diverse as possible, we quantized each real-valued state into 10 bins. 




\subsection{Action Space}

There are 3 different action spaces for three different agents. These agents are called the \textit{skills agent}, the agent that learns skill policies, \textit{low-level Deep Q-Network (DQN) agent}, which is used for comparison purposes, and finally the proposed \textit{high-level DQN agent}, which is trained with hierarchical reinforcement learning and uses skill policies as low-level policies (see Section~\ref{sec:HRL-USD}). We explain the training of these agents in detail in the following sections. In this section, we provide the action spaces they use. 

\begin{itemize}
    \item Skills Agent: Skills agent has two action selections. The first is $a$, the acceleration, which is a real number in the range $ [-a_{max}, a_{max}]$, where $a_{max}=4.5m/s^2$. The second is $l_p$, the lane change probability, that takes values in the range $[-0.1,1.1]$. 
    \item Low-Level DQN Agent: This agent selects one of the following actions, where $Exp[\lambda]$ is defined as the exponential distribution with rate parameter $\lambda = 0.75$.
    \begin{itemize}
        \item \textbf{Maintain:} 
        Acceleration $a$ is sampled from a Laplace distribution with µ = 0, and b = 0.1, in the interval [-0.25m/s$^2$, 0.25m/s$^2$]. Lane change probability $l_p$ is set to 0.
        \item \textbf{Accelerate:} 
        Parameter $a_{act}$ is sampled from $Exp[\lambda]$, and then used to set $a$ to $\min \{ 0.25 + a_{act}, 2 \}\text{m/s}^2$. $l_p$ is set to 0.
        \item \textbf{Decelerate:} 
        $a_{act}$ is sampled from $Exp[\lambda]$, and $a$ is set to $\max \{ -0.25 - a_{act}, -2 \}\text{m/s}^2$. $l_p$ is set to 0.
        \item \textbf{Hard-Accelerate:} 
        $a_{act}$ is sampled from $Exp[\lambda]$, and $a$ is set to $\min \{ 2 + a_{act}, 3 \}\text{m/s}^2$. $l_p$ is set to 0.
        \item \textbf{Hard-Decelerate:} 
         $a_{act}$ is sampled from $Exp[\lambda]$, and $a$ is set to $\max \{ -2 - a_{act}, -4.5 \}\text{m/s}^2$. $l_p$ is set to 0.
        \item \textbf{Merge:} 
        $a$ is set to 0, and $l_p$ is set to 1.
    \end{itemize}
    \item High-Level DQN Agent: This agent chooses a skill index $i \in \{ 1, 2, ..., n_{skills} \}$ as an action, and $i$ is translated in the skill $z_i$. Then, the agent reduces to a skills agent (see the first bullet point) and the acceleration $a$ and lane change probability $l_p$ actions are sampled from the corresponding skill policy $\pi_{z_i}(s)$.
\end{itemize}

\subsection{Environment Vehicles}
The vehicles in the environment have the same observation space with the ego agent but they select actions using Algorithm~1. In this algorithm, action definitions are the same as the ones given in Section III-E. TTC is the time to collision, calculated as $d_{rel}/v_{rel}$, which are defined in Section III-D, if $v_{rel}<0$, and is set to TTCd+1, otherwise. TTCd=5s and TTChd=3s are the time limits for using the actions Decelerate and Hard-Decelerate, respectively. All relative values mentioned here and in Algorithm~1 are referring to the vehicle in front. 
\begin{algorithm}[b]
    \caption{Environment Vehicle Algorithm}
    \begin{algorithmic}[1]
    \STATE $\text{action} \coloneqq \text{Maintain}$
    \IF{$\text{TTC} \leq \text{TTChd} \textbf{ or }  d_{rel} \leq d_{close}$} 
    \STATE $\text{action} \coloneqq \text{Hard-Decelerate}$
    \ELSIF{$\text{TTC} \leq \text{TTCd}$}
    \STATE $\text{action} \coloneqq \text{Decelerate}$
    \ELSIF{$v_{agent} \leq v_{nom}$}
    \STATE $\text{action} \coloneqq \text{Accelerate}$
    \ENDIF
    \end{algorithmic} 
\end{algorithm}

\subsection{Reward Function}

Reward function $r$ is the representation of a driver's preferences. In this study, it is defined as
\begin{equation}
    r = c \cdot w_c + h \cdot w_h + m \cdot w_m + n \cdot w_{nm},
\end{equation}
where $w$ terms are the corresponding weights of each feature. Features are defined below. 

\begin{itemize}
    \item \textbf{c:} Collision parameter. Takes the value $-1$ if the ego vehicle collides with another vehicle or if it reaches the end of the merging region without merging. The parameter gets the value 0, otherwise.
    
    \item \textbf{h:} Headway parameter. Calculated as 
    \begin{equation}
        h = \begin{cases} 
            -1 & d_{front} < d_{close} 
            \\
            \frac{d_{front}-d_{nom}}{d_{nom}-d_{close}} & d_{close} \leq d_{front}< d_{nom} 
            \\
            0 & d_{nom} \leq d_{front},
        \end{cases}
    \end{equation}
    where $d_{nom} = 23.3$m is the mean headway distance in the dataset. $d_{close} = 3.9$m and $d_{far} = 29.34$m are defined using the mean of bottom 10\%, and top 10\% subsets of vehicles. Finally, $d_{front}$ is the relative distance of the vehicle in front of the ego agent.
    
    \item \textbf{m:} Velocity parameter. Calculated as
    \begin{equation}
        m= \begin{cases} 
          \frac{v_{agent} - v_{nom}}{v_{nom}}  &  v_{agent} \leq v_{nom} \\
          \frac{v_{nom}-v_{agent}}{v_{max}-v_{nom}}  &  v_{agent} > v_{nom},
        \end{cases}
    \end{equation}
    where $v_{nom}$ is the nominal velocity for the agent which is set to 9.01m/s, the mean velocity observed in the dataset (see Section~\ref{sec:dataparams}).
    
    \item \textbf{nm:} 
    “Not Merging” parameter. Equals to -1 when the agent is on the ramp, 0 otherwise. This parameter discourages the ego agent to keep driving on the ramp without merging.
\end{itemize}
%

\section{Training and Simulation Results}

\subsection{Model Initialization}
\label{sec:init}
All network weights are initialized with Xavier normal initialization \cite{Xavier Normal} with a scaling of 1.

\subsubsection{Skills Agent}

Skills agent consists of a policy network, value network, discriminator network and two Q-value networks. Each network has,
\begin{itemize}
    \item two fully connected hidden layers with 64 neurons and leaky-ReLU activation function with a slope of 1/100. 
    \item a fully connected output layer.
\end{itemize}
An empty replay buffer $M_{skill}$ with size $n_{buffer}\coloneqq10000$ is initialized. Number of skills, $n_{skills}$, is set to 10.

\subsubsection{DQN Agents}

Both the low-level and high-level DQN networks are implementations of DQN \cite{DQN}. Q-value networks for both agents have

\begin{itemize}
    \item three fully connected hidden layers with 64 neurons and leaky-ReLU activation function with a slope of 1/100. 
    \item a fully connected output layer.
\end{itemize}

Empty replay buffers $M_{low}$ and $M_{high}$ with size $10^6$ are initialized for the low-level and high-level agents respectively. The exploration rates, denoted as $\varepsilon_{low}$ and $\varepsilon_{high}$, begin at 1 and linearly decrease until they reach 0.05. This reduction takes place over 35\% of the total training time, equivalent to 70 seconds in a 200-second training session. The model undergoes updates in the form of 8 gradient steps at every 16 training steps, adhering to a learning rate of $0.9\times10^{-3}$ and employing a batch size of 512.


\subsection{Training Skill Policies with Controlled Randomization}

After the initialization of the networks, the skill policies are trained for 10000 episodes, following the algorithm in \cite{DIAYN}. To enhance the efficacy of the training method, we introduce controlled randomness: After a skill policy selects the actions, the ego vehicle is updated with the replacement actions $a^{train}=a+a^{rand}$ and $l_p^{train}=l_p+l_p^{rand}$, where $a^{rand}$ is sampled from normal distribution with mean 0 and standard deviation 1, and $l_p^{rand}$ is sampled from uniform distribution in the range [-0.1,0.1]. The controlled randomness elevates the quality and diversity of the acquired skills, which increases their utility as low-level policies. 

The skill policies obtained at the end of this process vary in complexity. Some policies make the vehicle go to a specific coordinate and come to a stop, while another one makes it keep speeding up until a collision with another vehicle. It is noted that a skill that permits a collision can be useful if the high-level agent learns to pair this skill with another one, where the combined skill-set provides a desired trajectory without collision.

Two examples of the skill policies are given in figures 3 and 4. In these figures, the red vehicle is the ego vehicle, and blue vehicles constitute the environment. The red vertical line in figure 3 is the coordinate where the ego vehicle makes a lane change. The top frame is the starting frame and each following row represents the progression of the scenario, from top to bottom.

\begin{figure}
    \centering  \includegraphics[width=0.5\textwidth]{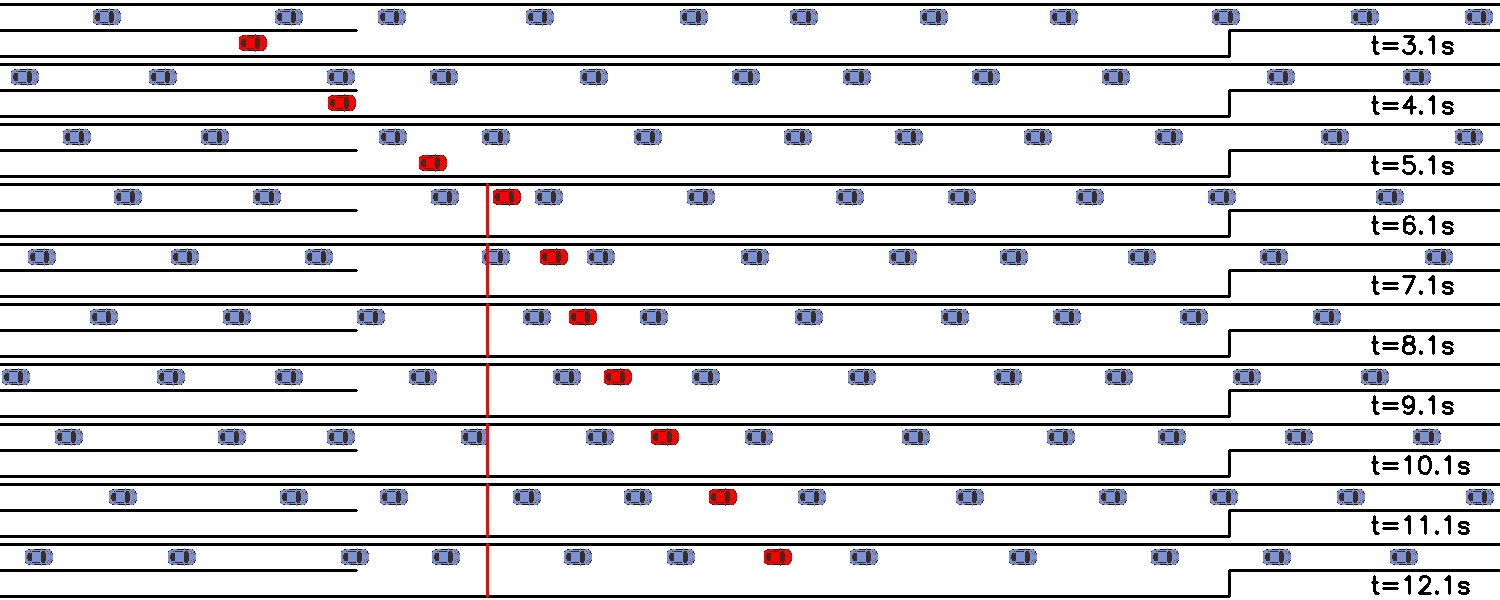}
    \caption{Example skill 1.}
    \label{fig:adsfasdgd}
\end{figure}
\begin{figure}
    \centering  \includegraphics[width=0.5\textwidth]{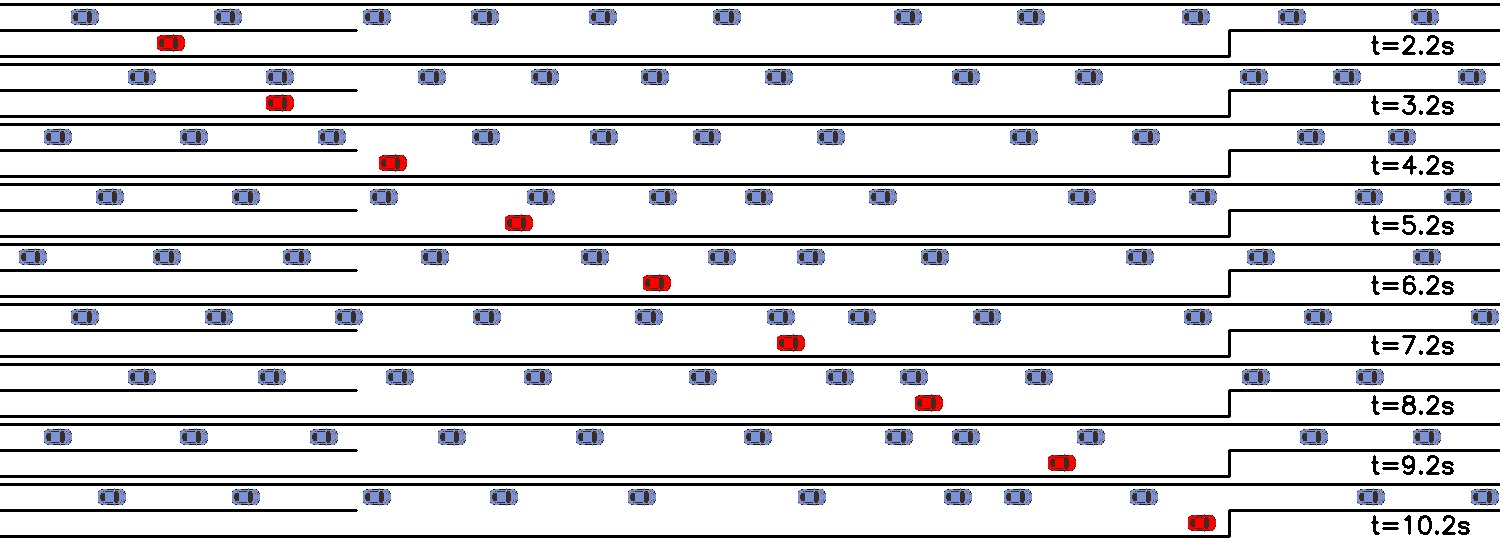}
    \caption{Example skill 2.}
\end{figure}

In figure 3, the ego vehicle merges in between two vehicles and stays in the middle of them until the end of the road. Such a skill can be utilized by the higher level agent for merging, assuming that the relative velocity and distance values are appropriate. In figure 4, the ego vehicle keeps accelerating until it crashes into the end of the merging region. Even though this might look like a counterproductive strategy, it could be useful when paired with other skills. For example, it could be used to quickly pass by a vehicle on its left, which would then be followed by a merging skill.

\subsection{Training Low and High-Level DQN Policies}

\begin{figure}[t]
    \centering
    \includegraphics[width=0.5\textwidth]{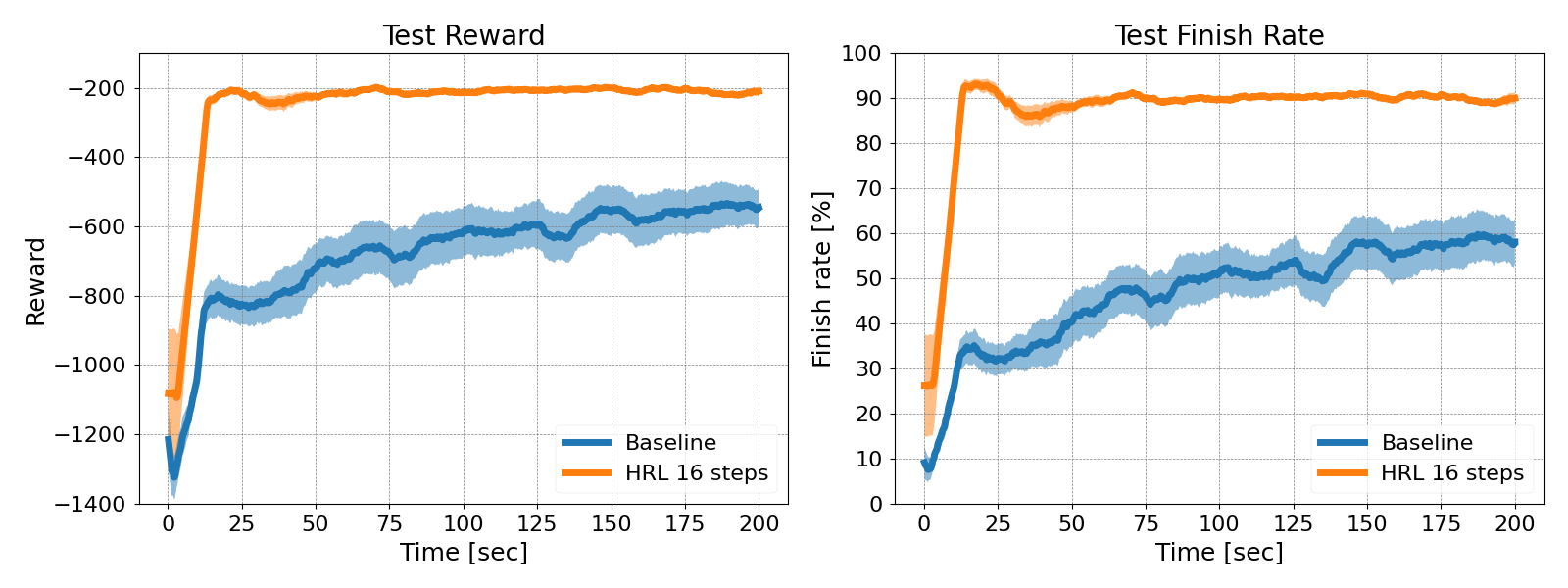}
    \caption{HRL and Baseline Model's Reward and Finish Rates vs. Training Time}
    \label{fig:results}
\end{figure}

Networks, exploration rates and memory buffers for DQN agents are initialized as explained in Section~\ref{sec:init}. Both the low-level and high-level DQN agents are trained for 200 seconds. The algorithm used to train the proposed high-level DQN agent is similar to the low-level one \cite{DQN} except for some key differences: The high-level agent selects one skill for every $n_{step}=16$ steps and that skill is applied throughout this duration. The selected skill $z$, the initial state $s$, the state after $n_{step}$ steps, $s_{fin}'$, and the cumulative reward observed during $n_{step}$ steps, $r_{cm}$ constitute one \textit{experience sample} for the buffer. If the episode terminates before $n_{step}$ steps are completed, $s_{fin}'$ and $r_{cm}$ are calculated using the final state before termination and the total reward observed so far.


In figure 5, the average rewards and finish rates of baseline (Low-level DQN) and the proposed HRL (High-level DQN using skills) models are displayed. Here, ``finish rate" refers to the percentage of episodes the ego agent reaches the end of the road without any collisions. It is noted that the HRL model selects one skill in every 16 steps to be applied for that duration. This interval is necessary to take advantage of skills, as skills represent trajectories that lead to a state subset, and choosing a skill at each step would be redundant and computationally inefficient. 

To obtain the graphs in figure 5, 10 test episodes with exploration rate set to zero is run every 0.2 seconds and the averages of the episodic rewards and finish rates are stored. Then, the running average of the last 15 seconds at each point is calculated. To account for the randomness of the initial model weights and of the environment, this process was repeated 10 times. Figure 5 shows the mean of these 10 runs together with the accompanying error in the mean bands.

It is observed that the skill-based model achieves higher average rewards and finish rates, while also converging with less training time than the baseline model. This is expected since skill-based RL needs to learn only how to use the skills, while baseline RL must learn to navigate the whole state space. Furthermore, the training process of the skill-based model appears to be more stable, as denoted by the smaller error bands. It is noted that since the proposed HRL selects new skills at every 16 steps, an exploratory move during training lasts 16 steps, which negatively affects the rewards. That is why although a non-zero exploration rate is used during training, the rewards are calculated using tests with zero exploration rate in figure 5.  

\section{CONCLUSIONS}

In this study, a hierarchical learning model that uses skills as actions is proposed for obtaining driving strategies. These skills can be obtained without a predefined reward function. This allows the skills to be reused in scenarios with differing reward functions to generate driving strategies with divergent behavior. Our simulation results show that, skill-based policies provide higher performance, and they can be obtained with less training compared to baseline reinforcement learning methods.

It can be argued that selecting skills as actions to be applied for a set amount of time instead of selecting primitive actions at each time step aligns more closely with human behavior. Humans have a delayed reaction time and take time to analyze the environment before deciding what to do next, which corresponds to the time the hierarchical model waits between each skill selection. Furthermore, humans generally do not consciously deliberate about specific motor skills, such as turning the wheel, and instead make high-level plans and execute the low-level actions automatically. These provide enough motivation to consider the proposed hierarchical driving model with skills as a promising direction for human behavior modeling in future transportation studies.



\begin{thebibliography}{00}

\bibitem{Dota}
 OpenAI et al., “Dota 2 with Large Scale Deep Reinforcement Learning,” arXiv:1912.06680 [cs, stat], Dec. 2019.
\bibitem{Atari}
J. Schrittwieser et al., “Mastering Atari, Go, chess and shogi by planning with a learned model,” \textit{Nature}, vol. 588, no. 7839, pp. 604–609, Dec. 2020.
\bibitem{Silver Go}
D. Silver et al., “Mastering the game of Go without human knowledge,” \textit{Nature}, vol. 550, no. 7676, pp. 354–359, Oct. 2017.
\bibitem{Silver All}
D. Silver et al., “A general reinforcement learning algorithm that masters chess, shogi, and Go through self-play,” \textit{Science}, vol. 362, no. 6419, pp. 1140–1144, Dec. 2018.
\bibitem{RL AV 1}
A. E. Sallab, M. Abdou, E. Perot, and S. Yogamani, “Deep Reinforcement Learning framework for Autonomous Driving,” \textit{Electronic Imaging}, vol. 2017, no. 19, pp. 70–76, Jan. 2017.
\bibitem{RL AV 2}
P. Wang, C.-Y. Chan, and A. de L. Fortelle, “A Reinforcement Learning Based Approach for Automated Lane Change Maneuvers,” \textit{IEEE Intelligent Vehicles Symposium (IV)}, Jun. 2018.
\bibitem{RL AV 3}
J. Chen, B. Yuan, and M. Tomizuka, “Model-free Deep Reinforcement Learning for Urban Autonomous Driving,” \textit{IEEE Intelligent Transportation Systems Conference (ITSC)}, Oct. 2019.
\bibitem{RL AV 4}
S. Kardell and M. Kuosku, “Autonomous vehicle control via deep reinforcement learning,” MSc Thesis, \textit{Chalmers University of Technology}, 2017. Accessed: Mar. 31, 2023. [Online]. Available: https://publications.lib.chalmers.se/records/fulltext/252902/252902.pdf
\bibitem{RL AV 5}
C. Li and K. Czarnecki, “Urban driving with multi-objective deep
reinforcement learning,” \textit{18th International Con
ference on Autonomous Agents and MultiAgent Systems}, pp. 359–367, 2019.
\bibitem{RL AV Survey}
B. R. Kiran et al., “Deep Reinforcement Learning for Autonomous Driving: A Survey,” \textit{IEEE Transactions on Intelligent Transportation Systems}, pp. 1–18, 2021.
\bibitem{Reward Shaping First}
H. Sowerby, Z. Zhou, and M. L. Littman, “Designing Rewards for Fast Learning,” arXiv:2205.15400 [cs], May 2022.
\bibitem{Reward Shaping Invariance}
T. Okudo and S. Yamada, “Subgoal-based Reward Shaping to Improve Efficiency in Reinforcement Learning,” \textit{IEEE Access}, pp. 1–1, 2021.
\bibitem{Macro HRL} 
J. Randlov, “Learning Macro-Actions in Reinforcement Learning,” \textit{Advances in Neural Information Processing Systems}, vol. 11, 1999.
\bibitem{Feudal HRL}
A. S. Vezhnevets et al., “FeUdal Networks for Hierarchical Reinforcement Learning,” arXiv:1703.01161 [cs], Mar. 2017.
\bibitem{Option-Critic}
P.-L. Bacon, J. Harb, and D. Precup, “The Option-Critic Architecture,” \textit{Proceedings of the AAAI Conference on Artificial Intelligence}, vol. 31, no. 1, Feb. 2017.
\bibitem{Act to Reason}
Koprulu, C., Yildiz, Y., “Act to Reason: A Dynamic Game Theoretical Driving Model for Highway Merging Applications,” \textit{IEEE Conference on Control Technology and Applications}, 2021.
\bibitem{Kickstarting}
S. Schmitt et al., “Kickstarting Deep Reinforcement Learning,” arXiv:1803.03835 [cs], Mar. 2018.
\bibitem{Hierarchical Kickstarting}
M. Matthews, M. Samvelyan, J. Parker-Holder, E. Grefenstette, and T. Rocktäschel, “Hierarchical Kickstarting for Skill Transfer in Reinforcement Learning,” arXiv:2207.11584 [cs], Aug. 2022.
\bibitem{HRL Book} 
A. van den Bosch, B. Hengst, J. Lloyd, R. Miikkulainen, and H. Blockeel, “Hierarchical Reinforcement Learning,” \textit{Encyclopedia of Machine Learning}, pp. 495–502, 2011.
\bibitem{HRL Survey}
S. Pateria, B. Subagdja, A.-H. Tan, and C. Quek, “Hierarchical Reinforcement Learning: A Comprehensive Survey,” \textit{ACM Computing Surveys}, vol. 54, no. 5, pp. 1–35, Sep. 2021.
\bibitem{Active Pre-Training}
H. Liu and P. Abbeel, “Behavior From the Void: Unsupervised Active Pre-Training,” arXiv:2103.04551 [cs], Oct. 2021.
\bibitem{Marginal Matching}
L. Lee, B. Eysenbach, E. Parisotto, E. Xing, S. Levine, and R. Salakhutdinov, “Efficient Exploration via State Marginal Matching,” arXiv:1906.05274 [cs, stat], Feb. 2020.
\bibitem{Curiosity}
D. Pathak, P. Agrawal, A. A. Efros, and T. Darrell, “Curiosity-driven Exploration by Self-supervised Prediction,” arXiv:1705.05363 [cs, stat], May 2017.
\bibitem{Proto RL}
D. Yarats, R. Fergus, A. Lazaric, and L. Pinto, “Reinforcement Learning with Prototypical Representations,” arXiv:2102.11271 [cs], Jul. 2021.
\bibitem{DADS} 
A. Sharma, S. Gu, S. Levine, V. Kumar, and K. Hausman, “Dynamics-Aware Unsupervised Discovery of Skills,” arXiv:1907.01657 [cs, stat], Feb. 2020.
\bibitem{DIAYN} 
B. Eysenbach, A. Gupta, J. Ibarz, and S. Levine, “Diversity is All You Need: Learning Skills without a Reward Function,” arXiv:1802.06070 [cs], Oct. 2018.
\bibitem{DQN}
F. Tan, P. Yan, and X. Guan, “Deep Reinforcement Learning: From Q-Learning to Deep Q-Learning,” \textit{Neural Information Processing}, pp. 475–483, 2017.
\bibitem{SAC}
T. Haarnoja, A. Zhou, P. Abbeel, and S. Levine, “Soft Actor-Critic: Off-Policy Maximum Entropy Deep Reinforcement Learning with a Stochastic Actor,” arXiv:1801.01290 [cs, stat], Aug. 2018.
\bibitem{NGSIM}
J. Halkias and J. Colyar, “Interstate 80 Freeway Dataset, FHWA-HRT-06-137,” www.fhwa.dot.gov, Dec. 2006. https://www.fhwa.dot.gov/publications/research/operations/06137/ (accessed Mar. 31, 2023).
\bibitem{NGSIM Reconstructed}
M. Montanino and V. Punzo, “Trajectory data reconstruction and simulation-based validation against macroscopic traffic patterns,” \textit{Transportation Research Part B: Methodological}, vol. 80, pp. 82–106, Oct. 2015.
\bibitem{Xavier Normal}
X. Glorot, Y. Bengio, "Understanding the difficulty of training deep feedforward neural networks," \textit{International Conference on Artificial Intelligence and Statistics}, pp. 249-256, 2010.
\end{thebibliography}
\end{document}